# Smartphone Transportation Mode Recognition Using a Hierarchical Machine Learning Classifier and Pooled Features from Time and Frequency Domains

Huthaifa I. Ashqar, Mohammed H. Almannaa, Mohammed Elhenawy, Hesham A. Rakha, *IEEE*, and Leanna House

*Abstract*—The paper develops a novel two-layer hierarchical classifier that increases the accuracy of traditional transportation mode classification algorithms. The study also enhances classification accuracy by extracting new frequency domain features. Many researchers have obtained these features from Global Positioning System (GPS) data; however, this data was excluded in our study, as the system use might deplete the smartphone's battery and signals may be lost in some areas. Our proposed two-layer framework differs from previous classification attempts in three distinct ways: (1) the outputs of the two layers are combined using Bayes' rule to choose the transportation mode with the largest posterior probability; (2) the proposed framework combines the new extracted features with traditionally used time domain features to create a pool of features; (3) a different subset of extracted features is used in each layer based on the classified modes. Several machine learning techniques were used, including k-nearest neighbor, classification and regression tree, support vector machine, random forest, and a heterogeneous framework of random forest and support vector machine. Results show that the classification accuracy of the proposed framework outperforms traditional approaches. Transforming the time domain features to the frequency domain also adds new features in a new space and provides more control on the loss of information. Consequently, combining the time domain and the frequency domain features in a large pool and then choosing the best subset results in higher accuracy than using either domain alone. The proposed two-layer classifier obtained a maximum classification accuracy of 97.02%.

## I. INTRODUCTION

The application of smartphones to data collection has recently attracted researchers' attention. Smartphone applications (apps) have been developed and effectively used to collect data from smartphones in many sectors. In the transportation sector, researchers can use smartphones to track and obtain information such as speed, acceleration, and the rotation vector from the built-in Global Positioning System (GPS), accelerometer, and gyroscope sensors [1]. These data can be used to recognize the user's transportation mode, which can be then be utilized in a number of different applications, as shown in Table I.

TABLE I
TRANSPORTATION MODE DETECTION APPLICATIONS [2]

| Application | Description |
|---|---|
| Transportation Planning | Instead of using traditional approaches such as questionnaires, travel diaries, and telephone interviews [3, 4], the transportation mode information can be automatically obtained through smartphone sensors. |
| Safety | Knowing the transportation mode can help in developing safety applications. For example, violation prediction models have been studied for passenger cars and bicycles [5, 6]. |
| Environment | Physical activities, health, and calories burned, and carbon footprint associated with each transportation mode can be obtained when the mode information is available [7]. |
| Information Provision | Traveler information can be provided based on the transportation mode [4, 8]. |

In this study, we investigated the possibility of improving the overall accuracy of transportation mode detection by proposing a new hierarchical framework classifier and by looking for a new features set. This paper makes two major contributions to the body of transportation research. First, it proposes a two-layer hierarchical framework in which a) the first layer contains one multi-classifier using the data set of the five transportation modes, and b) the second layer consists of 10 binary classifiers, each of which is specialized in only one pair of modes, and uses a features subset that discriminates between this pair. Second, new frequency domain features were extracted and pooled with the traditionally-used time domain features.

Following the introduction, this paper is organized into six sections. First, the approaches, features, and machine learning techniques of previous studies are reviewed. Next, the data set and the extracted features are described. Third, background is presented on the machine learning techniques applied in this study. Next, the proposed framework is presented. In the fifth section, details are provided on the data analysis used to detect different transportation modes. Finally, the paper concludes with a summary of new insights and recommendations for future transportation mode recognition research.

## II. RELATED WORK

Researchers have developed several approaches to discriminate between transportation modes effectively using mobile phones [9,10] or visual tracking [11]. Machine learning techniques have been used extensively to build detection models and have shown high accuracy in determining transportation modes. Supervised learning methods such as K-Nearest Neighbor (KNN) [12], Support Vector Machines (SVMs) [7, 13-17], Decision Trees [3, 4, 7, 8, 14, 18], and Random Forests (RFs) [12], have all been employed in various studies.

These studies have obtained different classifying accuracies. There are several factors that affect the accuracy of detecting transportation modes, such as the *monitoring period* (positive association)*, number of modes* (negative association)*, data sources, motorized classes,* and *sensor positioning* [2, 12].

However, one of the most critical factors that affects the accuracy of mode detection is the machine learning framework classifier. The framework that usually uses one layer of classification algorithm as in [4, 7] could be refered to as traditional framework; whereas the hierarchal framework uses more than one layer of classification algorithm.

An additional important consideration is the domain of the extracted features. Features are generally extracted from two different domains: (1) the time domain, features of which have been used widely in many studies [4, 12, 15, 16, 19, 20]; and (2) the frequency domain, features of which have been used in some studies [7, 15]. Both methods have achieved a significant, high accuracy. Table III summarizes the obtained accuracies and factors for some of the aforementioned studies. Note that no direct comparison can be made between the studies listed in Table III because the factors considered and the data sets used varied from study to study.

In this study, we will mainly focus in the effect of two factors on the accuracy of transportation mode detection; namely the framework classifier and extracted features.

Most of the proposed methods in the most recent studies rely on the using of the GPS data, which do not take into account the limitations of GPS information. GPS service is not available or may be lost in some areas, which results in inaccurate position information. Moreover, the GPS system use might deplete the smartphone's battery. Thus, this paper focuses on proposing a new detection framework using machine learning techniques and extract new features based on data obtained from smartphone sensors including accelerometer, gyroscope, and rotation vector, without GPS data.

## III. DATA SET

### A. Data Collection

The data set used is available at the Virginia Tech Transportation Institute (VTTI) and was collected by Jahangiri and Rakha [12] using a smartphone app (two devices were used: a Galaxy Nexus and a Nexus 4) [12]. The app was provided to 10 travelers who work at VTTI to collect data for five different modes: driving a passenger car, bicycling, taking a bus, running, and walking. The data were collected from GPS, accelerometer, gyroscope, and rotation vector sensors and stored on the devices at the application's highest possible frequency. Data collection was conducted on different workdays (Monday through Friday) and during working hours (8:00 a.m. to 6:00 p.m.). Several factors were considered to collect realistic data reflecting natural behaviors. No specific requirement was applied in terms of sensor positioning other than carrying the smartphone in different positions that they normally do, to make sure the data collection is less dependent on the sensor positioning. The data were collected on different road types with different speed limits in Blacksburg, Virginia, and some epochs may reflect traffic jam conditions occurring in real-world conditions. The collection of thirty minutes of data over the course of the study for each mode per person was considered sufficient.

For the purpose of comparing data with data from previous studies [2, 12], the extracted features were considered to have a meaningful relationship with different transportation modes. Furthermore, features that might be extracted from the absolute values of the rotation vector sensor were excluded.

Additionally, in order to allow this framework to be implemented in cases where no GPS data were available, features that might be extracted from GPS data were also excluded.

*B. Time Domain Features*

From the time window $t$, time domain features were created by applying the measures in Table II. These measures were applied using the measurements of the data array $x_i^t$ and its derivative $\dot{x}_i^t$ for the $i^{th}$ feature from time window $t$. This resulted in 165 time domain features: out of the 18 measures presented in in Table II, all the 18 measures were applied to accelerometer and gyroscope sensor values; 7 measures were applied to rotation vector sensor values; 16 measures were applied to the summation values from accelerometer and gyroscope sensors; 4 measures were applied to the summation values from rotation vector sensor. As a result, the total number of features reached $18(6) + 7(3) + 16(2) + 4(1) = 165$ features [12].

TABLE II
MEASUREMENTS OF TIME DOMAIN FEATURES [12].

| No. | Measure | No | Measure |
|---|---|---|---|
| 1 | $mean(x_i^t)$ | 10 | $spectralEntropy(x_i^t)$ |
| 2 | $max(x_i^t)$ | 11 | $mean(\dot{x}_i^t)$ |
| 3 | $min(x_i^t)$ | 12 | $max(\dot{x}_i^t)$ |
| 4 | $variance(x_i^t)$ | 13 | $min(\dot{x}_i^t)$ |
| 5 | $standard\ deviation(x_i^t)$ | 14 | $variance(\dot{x}_i^t)$ |
| 6 | $range(x_i^t)$ | 15 | $standard\ deviation(\dot{x}_i^t)$ |
| 7 | $Interquartile\ range(x_i^t)$ | 16 | $range(\dot{x}_i^t)$ |
| 8 | $signChange(x_i^t)$ | 17 | $Interquartile\ range(\dot{x}_i^t)$ |
| 9 | $energy(x_i^t)$ | 18 | $signChange(\dot{x}_i^t)$ |

*C. Frequency Domain Features*

Jahangiri and Rakha [12] collected readings from the mobile sensors at a frequency of almost 25 Hz. Because the output samples of the sensors were not synchronized, the authors implemented a linear interpolation to build continuous signals from the discrete samples. Consequently, they sampled the constructed sensor signals at 100 Hz and divided the output of each sensor in each direction $(x, y, and\ z)$ into non-overlapping windows of 1-s width. Finally, the features used for mode recognition were extracted from each window. These features were mainly traditional statistics such as mean, minimum, and maximum. The use of these features achieved a good accuracy in mode recognition.

However, some information loss was expected because of the usage of the summary statistics. Summary statistics consist of some descriptive statistics analysis for variability, center tendency, and distribution, such as mean, range, and variance. Summary statistics occasionally fail to detect the correlations, and extract optimal information and define probabilities [21, 22].

Since each window is considered as a signal in the time domain, we transferred each signal into the frequency domain using the short-time Fourier transform. Fourier transform converts the time function into a sum of sine waves of different frequencies, each of which represents a frequency component. The spectrum of frequency components is the frequency domain representation of the signal. Further, the component frequencies, which are spread across the frequency spectrum, are represented as peaks in the frequency domain. These peaks represent the most dominant frequencies in the signal. However, a frequency domain can also include information on the phase shift that could be applied to each sinusoid in order to be able to recombine the frequency components to recover the original time signal. In that sense, after transforming the time domain signal to the frequency domain and neglecting the phase information, we visually inspected the resultant spectrum and found that most of the information was provided by the first 20 resulted components, which means the highest 20 magnitude of that signal in the frequency domain.

In this study, we used the magnitude of these 20 components as the new frequency independent features. Transforming the time domain into the frequency domain not only adds new transferred features from an original space (i.e., time) to a new space (i.e., frequency), but also imposes more control on the loss of information. While the time domain represents the signal changes over time, the frequency domain add to the time domain features: how much of the signal lies within each given frequency band over a range of frequencies. As a result, some of the expected loss in the information about signal changes in the time domain features (because of the usage of the summary statistics) might be substituted by extracting features from the frequency domain. This process resulted in the addition of another 180 features extracted from the frequency domain to the data set (i.e., 345 features pooled in total).

TABLE III.
SUMMARY OF SOME PAST STUDIES [2].

| Accuracy (%) | Features Domain | Machine Learning Framework | Monitoring Period | No. of Modes | Data Sources | More than One Motorized Mode? | Sensor Positioning | Data Set | Study |
|---|---|---|---|---|---|---|---|---|---|
| 97.31 | Time | Traditional | 4 s | 3 | Accelerometer | Yes | No requirements | Not mentioned | [16] |
| 93.88 | Frequency | Traditional | 5 s, 50% overlap | 6 | Accelerometer | Yes/No | Participants were asked to keep their device in the pocket of their non-dominant hip | Collected from 4 participants | [15] |
| 93.60 | Time and frequency | Traditional | 1 s | 5 | Accelerometer GPS | No | No requirements | Collected from 16 participants | [7] |
| 93.50 | Time | Traditional | 30 s | 6 | GPS, GIS[a] maps | Yes | No requirements | Collected from 6 participants | [4] |
| 95.10 | Time | Traditional | 1 s | 5 | Accelerometer, gyroscope, rotation vector | Yes | No requirements | Collected from 10 participants | [12] |
| 91.60 | Time | Traditional | Entire trip | 11 | GPS, GIS maps | Yes | No requirements | Two different data sets, one of which included 1,000 participants | [20] |
| 96.32 | Time | Hierarchical | 1 s | 5 | Accelerometer, gyroscope, rotation vector | Yes | No requirements | Collected from 10 participants | [2] |

[a] GIS: Geographic Information System

## IV. METHODS

This section describes the feature selection algorithm and the machine learning classifiers used in the proposed hierarchical framework.

### A. K-Nearest Neighbor (KNN)

KNN is a common algorithm in supervised learning that classifies the data points based on the K nearest points. K is a user parameter that can be determined using different techniques. The test observation (i.e., $y_j^{test}$) is classified by taking the majority vote of the classes of the K nearest points (i.e., $y_j^{train}$), as shown in Equation (1) [23].

$$y_j^{test} = \frac{1}{K} \sum_{X_j^{train} \in N_K} y_j^{train} \qquad (1)$$

where, $y_j^{test}$ is the class of the testing data; $y_j^{train}$ is the class of the training data; $X_j^{train}$ is the testing data; and $K$ is the number of classes.

### B. Classification and Regression Tree (CART)

The CART algorithm was introduced in the early 1980s by Olshen, and Stone [24]. This algorithm is a type of decision tree where each branch represents a binary variable. At each split, the CART algorithm trains the tree using a greedy algorithm. Different splits are tested, and the split with the lowest cost is chosen. After many splits, each branch will end up in a single output variable that is used to make a single prediction. The CART algorithm will stop splitting upon reaching a certain criterion. The two most common stopping criteria are setting a minimum count of the training instances assigned to each leaf and choosing a pruning level that produces the highest accuracy.

### C. Support Vector Machines (SVMs)

The SVM algorithm is a supervised learning technique that is used to classify the data by maximizing the gap between classes. The SVM algorithm attempts to find the hyperplane (i.e., splitter) that gives the largest minimum distance to the training data as given in Equation (2). The SVM tries to find the weight ($w$) that produces the largest margin around the hyperplane (see Equation (2)), while satisfying the two constraints (see Equations (3) and (4)) [25].

$$\min_{w,b,\xi} \left( \frac{1}{2} w^T w + C \sum_{n=1}^{N} \xi_n \right) \qquad (2)$$

subject to:

$$y_n(w^T\phi(x_n) + b) \geq 1 - \xi_n, n = 1, \ldots, N \quad (3)$$

$$\xi_n \geq 0, n = 1, \ldots, N \quad (4)$$

where,

| | |
|---|---|
| $w$ | Parameters to define the decision boundary between classes |
| $C$ | Penalty parameter |
| $\xi_n$ | Error parameter to denote margin violation |
| $b$ | Intercept associated with the hyperplanes |
| $\phi(x_n)$ | Function to transform data from X space into some Z space |
| $y_n$ | Target value for $n^{th}$ observation |

*D. Random Forest (RF)*

Breiman proposed RF as a new classification and regression technique in supervised learning [26]. The RF method randomly constructs a collection of decision trees in which each tree chooses a subset of features to grow, and the results are then obtained based on the majority votes from all trees. The number of decision trees and the selected features for each tree are user-defined parameters. The reason for choosing only a subset of features for each tree is to prevent the trees from being correlated. RF was applied in this study to select the best subset of features to be used in classification, as this technique offers several advantages. For example, it runs efficiently on large datasets and can handle many input features without the need to create extra dummy variables, and it ranks each feature's individual contribution in the model [26, 27].

## V. PROPOSED FRAMEWORK

As many features could be used to discriminate between transportation modes, we applied feature selection to choose the subset of features with the highest importance. The subset of selected features, which is used in the classifiers, depends on the classified modes. This implies that the subset of features selected to discriminate between all modes will be different from the subset of features selected to discriminate between only two modes. In this study, RF was used to select the best 100-feature subset for each classifying step. Selected features were scaled so that the feature values were normalized to be within the range of $[-1, 1]$.

Fig. 1 shows the importance of features in different ranks for all the modes combined and for different pairs of modes. The least important feature is ranked 0.1, the highest is ranked 2.2, and 0 when the features are not included. Fig. 1 also illustrates that the most important feature of one pair of modes may be different for other pairs and that its rank within mode pairs may also vary. It is noteworthy that the car-run and the car-walk pairs have lower scores as compared to the other (most of the features have a score of 0.5, which is shown in dark blue). It appears that the values of some selected features for pairs containing walk and run modes are more likely to overlap. RF ranks each feature's individual contribution in the model relatively, which means the overlap would affect the score of the individual features but not the overall classification accuracy using the entire subset of features. The overlap occurs between the features in some level of dimensionality and could be separable in the higher dimensions with high accuracy.

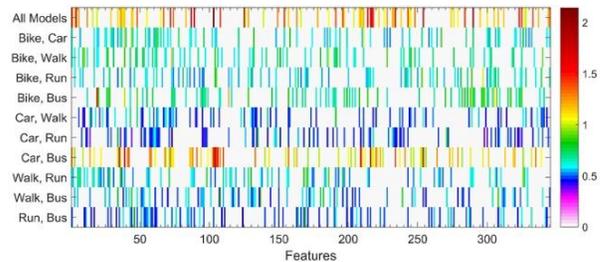

Fig. 1. Importance of features for different pairs of modes.

This study proposes a new approach to detect transportation modes. Two layers are applied as a hierarchical framework. The first layer consists of only a one multiclass classifier to discriminate between the five modes, and the second layer consists of a pool of 10 binary classifiers, which are used to discriminate between only two modes. The output of each classifier (in the first or the second layer) is the probability of each mode given the test data. The first layer is trained using the RF-selected 100 features to return the corresponding modes with the highest ($i$) and the second highest ($j$) probabilities ($M^{(1)} \in \{i, j\}$). These two modes are the candidates for input to the second layer. Each classifier in the second layer is trained using a different set of RF-selected 100 features, specialized to differentiate between only the two modes of interest, to return one mode of the highest probability ($M^{(2)} \in \{i\}$). Bayesian principles are used in this framework to combine the output of the two layers. In that sense, the transportation mode with the largest posterior probability is chosen, given that the output of the first layer is the prior probability and the output of the second layer is the likelihood.

In the first layer, the probability that $M^{(1)} \in \{i\}$ is the true mode ($T$) in a one-layer traditional framework (i.e., $p(M^{(1)} \in \{i\}|T)$) equals $P^{(1)}$. However, the probability that $M_i^{(1)}$ or $M_j^{(1)}$ (where $i \neq j$) is the true mode ($T$) in the two-layer proposed framework (i.e., $p(M^{(1)} \in \{i, j\}|T)$) equals $P^{(1)} + \Delta$.

Consequently, the proposed framework improves the potential to obtain the true mode by selecting two modes instead of only one in the first layer. The second layer consists of a pool of 10 binary classifiers ($k$). Thus, the probability that one mode, out of the two candidates from the first layer, is the true mode (i.e., $p(M^{(2)} \in \{i\})$) equals $c \sum_{k=1}^{10} P_k^{(2)}$. The constant $c$ equals $\frac{1}{k}$ if the data are assumed to be balanced. Consequently, the output of the framework can be formulated using the total law of probability, as shown in Equation (5):

$$p(M^{(1)} \in \{i,j\}, M^{(2)} \in \{i\}, T) =$$
$$\sum p(M^{(2)} \in \{i\}|M^{(1)} \in \{i,j\}) p(M^{(1)} \in \{i,j\}|T) p(T) \quad (5)$$

$$= \sum_{k=1}^{10} P_k^{(2)} \times (P^{(1)} + \Delta) \times c$$

$$= (P^{(1)} + \Delta)[1 - (1 - c \sum_{k=1}^{10} P_k^{(2)})]$$

substituting $(1 - c \sum_{k=1}^{10} P_k^{(2)})$ by the summation of the errors in the second layer ($c \sum_{k=1}^{10} e_k^{(2)}$);

$$= P^{(1)} + \Delta - (P^{(1)} + \Delta) c \sum_{k=1}^{10} e_k^{(2)}$$

In fact, this term $\Delta - (P^{(1)} + \Delta) c \sum_{k=1}^{10} e_k^{(2)}$ is the difference between the output of using a one-layer traditional framework and the output of using the two-layer proposed framework. Hence, if this term is greater than zero, then there is an additional amount to probability resulting from using a one-layer traditional framework. In that case, results from the proposed framework are better than the traditional framework. This can be formulated as shown in Equation (6).

$$\Delta - (P^{(1)} + \Delta) \sum_{k=1}^{10} e_k^{(2)} > 0 \quad (6)$$

$$\Delta > P^{(1)}(1 - c \sum_{k=1}^{10} P_k^{(2)}) / c \sum_{k=1}^{10} P_k^{(2)}$$

This implies that if $\Delta$ is greater than the term $[P^{(1)}(1 - c \sum_{k=1}^{10} P_k^{(2)}) / c \sum_{k=1}^{10} P_k^{(2)}]$, then the two-layer framework is beneficial. In order to examine that, we need to estimate 12 parameters from the data: $\Delta, P^{(1)}$, and $P_k^{(2)}$, where $k = 1, 2, \ldots, 10$. One reasonable method is to formulate the likelihood function of the classifier output as Bernoulli distribution because it is either one or zero, whereas the prior function for each the 12 parameters is formulated as a Beta distribution because it takes on any value between zero and one. This means that the prior domain is from zero to one [0,1]. Consequently, the problem can be viewed as a Beta-Bernoulli model:

$$f(y_{ij}|p_j) \sim Bernouli(p_j)$$

where $y_{ij} \in \{1,0\}$ is output $i$ for classifier $j$ and $p_j \in [0,1]$

$$f(p_j) \sim Beta(a, b)$$

where the values of constants $a$ and $b$ are chosen in which the knowledge of the prior is equal (i.e., $E[p_j] = 0.5$).

$$f(p_j|y_{j1}, y_{j1}, \ldots, y_{jN_j}) \propto$$
$$\left\{ \prod_{i=1}^{N_j} p_j^{y_{ji}} (1-p_j)^{1-y_{ji}} \right\} \frac{\Gamma(a+b)}{\Gamma(a)\Gamma(b)} p_j^{a-1} (1-p_j)^{b-1}$$

the above equation can be simplified as:

$$f(p_j|y_{j1}, y_{j1}, \ldots, y_{jN_j}) \propto$$
$$\frac{\Gamma(a+b)}{\Gamma(a)\Gamma(b)} \left\{ p_j^{\sum_{i=1}^{N_j} y_i + a - 1} (1-p_j)^{N_j - \sum_{i=1}^{N_j} y_i + b - 1} \right\}$$

by removing the constants $\frac{\Gamma(a+b)}{\Gamma(a)\Gamma(b)}$ from the above equation, the kernel of the posterior is a Beta distribution with the parameters shown as:

$$f(p_j|y_{j1}, y_{j1}, \ldots, y_{jN_j}) \sim Beta\left( \sum_{i=1}^{N_j} y_i + a, N_j - \sum_{i=1}^{N_j} y_i + b \right)$$

From the above equation we can estimate the expectation $E\left[ f(p_j|y_{j1}, y_{j1}, \ldots, y_{jN_j}) \right]$ of each parameter, as shown in Equation (7):

$$E\left[ f(p_j|y_{j1}, y_{j1}, \ldots, y_{jN_j}) \right] = \frac{\sum_{i=1}^{N_j} y_i + a}{a + b + N_j} \quad (7)$$

Each of the required parameters can be estimated using Equation (7). However, Equation (5), Equation (6), and the corresponding results are based on a two-layer framework. As the number of layers in the framework increases, more parameters are required to be estimated and the model will be relatively more complicated. In addition, adding layers to the framework would increase the computational time. Yet, this does not mean that adding layers is costly, so we recommend first estimating the parameters related to the number of layers one will choose, then decide upon that.

## VI. Data Analysis and Results

This section discusses the results of the machine learning techniques used in this study, which were developed in MATLAB.

### A. K-Nearest Neighbors Algorithm (KNN)

In this study, KNN was used to identify the mode from the five possible transportation modes in the first layer and the two modes in the second layer. The optimal $K$ was chosen after testing different numbers of $K$ versus the overall classification accuracy. To select the best model at each value of $K$, a 10-fold cross-validation was performed, and the average highest accuracy among the 10 folds was chosen. As shown in Fig. 2, a higher classification accuracy was achieved using the pooled features in the proposed hierarchical framework than using only the time domain features in the same framework. Additionally, using only the time domain features, the proposed hierarchical framework outperformed traditional KNN classification using pooled features. The optimal K was found to be 7, with a highest accuracy of 95.49%.

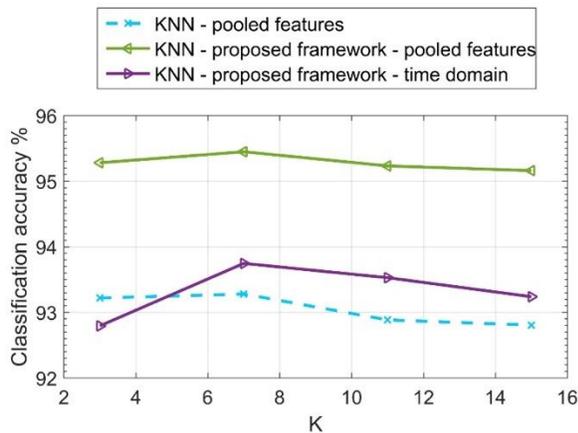

Fig. 2. Classification accuracy for KNN in different cases at different neighbors.

### B. Classification and Regression Tree (CART)

Ten folds for the cross-validation process were applied for each pruning level, ranging from two to 20, and the average was taken as a comparison value with other pruning levels. Fig. 3 provides a comparison between time domain features, frequency domain features, and pooled features for traditional CART and CART using the proposed framework under different pruning levels. The figure shows that the proposed framework using pooled features (compared to the same applied approach using traditional CART) produces the highest accuracy among all other cases (93.52%) at six pruning levels. Fig. 3 also shows that the classification accuracy of using only frequency domain features in the proposed framework approach (compared to the same approach using traditional CART) is lower than using the proposed approach and only time domain features.

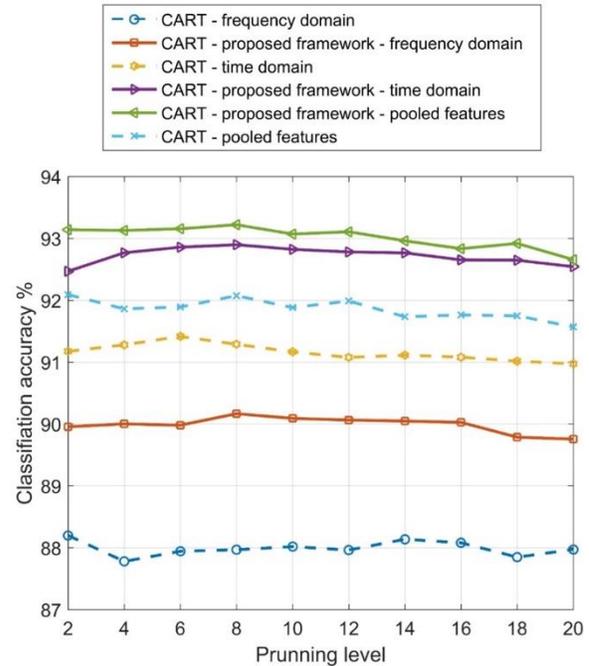

Fig. 3. Classification accuracy for CART in different cases at different pruning levels.

### C. Support Vector Machine (SVM)

SVM was applied in the proposed framework using time domain, frequency domain, and pooled features. A 10-fold cross-validation was applied to develop a single model. The results show that using pooled features improved the average overall classification accuracy from 96.10% to 97.00%. The overall accuracy for using only the frequency domain features was the lowest at 93.92%. Table IV presents the overall classification accuracy for the 10-fold testing applying the proposed SVM framework.

TABLE IV
OVERALL CLASSIFICATION ACCURACY FOR THE SVM USING TIME DOMAIN, FREQUENCY DOMAIN, AND POOLED FEATURES.

| Fold | Time domain features (%) | Frequency domain features (%) | Pooled features (%) |
|---|---|---|---|
| 1 | 96.04 | 93.78 | 97.12 |
| 2 | 96.32 | 93.65 | 97.31 |
| 3 | 95.88 | 92.90 | 96.88 |
| 4 | 96.10 | 93.04 | 97.32 |
| 5 | 95.98 | 94.01 | 96.76 |
| 6 | 96.02 | 94.79 | 96.81 |
| 7 | 96.38 | 93.62 | 96.98 |
| 8 | 95.71 | 94.57 | 96.93 |
| 9 | 96.32 | 94.52 | 97.01 |
| 10 | 96.25 | 94.28 | 96.91 |
| Average | 96.10 | 93.92 | 97.00 |

The confusion matrix applying SVM in the proposed framework using pooled features is given in Table V. The precision for run mode was the highest, and the precision for bus mode was the lowest. However, the recall was the lowest for run mode and highest for bike mode.

TABLE V
CONFUSION MATRIX FOR SVM USING POOLED FEATURES

|  |  | Actual |  |  |  |  |  |
|---|---|---|---|---|---|---|---|
|  |  | Bike | Car | Walk | Run | Bus | Precision |
| Predicted | Bike | 97.13 | 0.52 | 1.17 | 0.40 | 0.58 | 97.33 |
|  | Car | 0.66 | 93.57 | 0.16 | 0.13 | 3.06 | 95.88 |
|  | Walk | 0.92 | 0.08 | 93.59 | 0.92 | 0.29 | 97.68 |
|  | Run | 0.37 | 0.05 | 0.93 | 92.82 | 0.20 | 98.36 |
|  | Bus | 0.92 | 2.42 | 0.40 | 0.32 | 93.11 | 95.81 |
|  | Recall | 97.13 | 93.57 | 93.59 | 92.82 | 93.11 |  |

### D. Random Forest (RF)

We ran the RF with different numbers of trees to investigate the impact of the number of trees on the classification accuracy. A number of trees ranging from 200 to 400 was chosen, as the highest benefit was expected to be gained in this range according to previous studies (see more details in Elhenawy, Jahangiri, and Rakha; Jahangiri and Rakha [2, 12]). Applying RF in the proposed framework using pooled features resulted in the highest classification accuracy of 96.24% at 200 trees, as illustrated in Fig. 4. Fig. 4 also illustrates that applying RF using a traditional approach for classification with pooled features produces higher accuracy than the RF using the proposed classification framework with only time domain features.

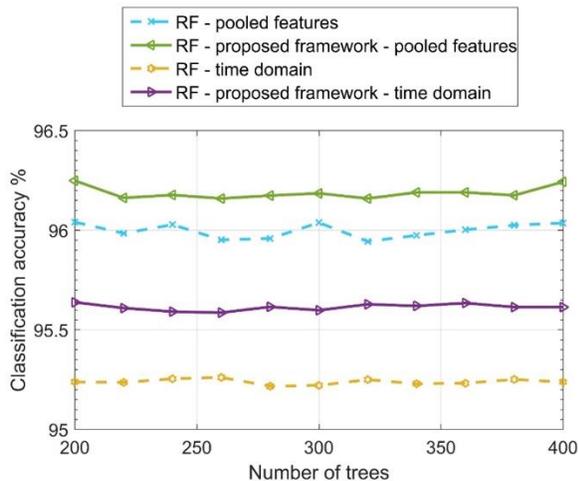

Fig. 4. Classification accuracy for RF in different cases at different number of trees.

A comparison between time domain, frequency domain, and pooled features was carried out using the RF method in the proposed framework, as shown in Table VI. The results demonstrate that using the pooled features improved the overall classification accuracy from 95.61% to 96.24%.

TABLE VI
OVERALL CLASSIFICATION ACCURACY FOR RF USING TIME DOMAIN, FREQUENCY DOMAIN, AND POOLED FEATURES

| Fold | Time domain features (%) | Frequency domain features (%) | Pooled features (%) |
|---|---|---|---|
| 1 | 95.95 | 94.15 | 96.35 |
| 2 | 95.73 | 94.34 | 96.59 |
| 3 | 95.61 | 93.91 | 96.07 |
| 4 | 95.37 | 94.02 | 96.22 |
| 5 | 95.51 | 93.85 | 96.24 |
| 6 | 95.56 | 94.09 | 96.30 |
| 7 | 95.67 | 93.82 | 96.23 |
| 8 | 95.78 | 93.64 | 96.13 |
| 9 | 95.49 | 94.18 | 96.39 |
| 10 | 95.47 | 93.78 | 95.88 |
| Average | 95.61 | 93.98 | 96.24 |

Table VII shows the confusion matrix for the RF proposed framework using pooled features. The run mode has the highest precision and the bus mode has the lowest precision.

TABLE VII
CONFUSION MATRIX FOR RF USING POOLED FEATURES

|  |  | Actual |  |  |  |  |  |
|---|---|---|---|---|---|---|---|
|  |  | Bike | Car | Walk | Run | Bus | Precision |
| Predicted | Bike | 94.63 | 0.40 | 2.59 | 0.05 | 0.94 | 95.96 |
|  | Car | 0.97 | 92.54 | 0.13 | 0.00 | 2.78 | 95.96 |
|  | Walk | 1.87 | 0.10 | 91.74 | 0.25 | 0.70 | 96.92 |
|  | Run | 0.75 | 0.05 | 1.47 | 90.39 | 0.57 | 96.96 |
|  | Bus | 1.78 | 2.43 | 0.13 | 0.00 | 91.67 | 95.48 |
|  | Recall | 94.63 | 92.54 | 91.74 | 90.39 | 91.67 |  |

### E. Heterogeneous Framework RF-SVM

We performed a heterogeneous framework in which the RF classifier was used in the first layer to classify all modes and a binary SVM classifier was applied in the second layer. The overall classification accuracy improved when using pooled features (from 96.32% to 97.02%) compared to when using only time domain features, as presented in Table VIII.

TABLE VIII
OVERALL CLASSIFICATION ACCURACY FOR RF-SVM USING TIME DOMAIN, FREQUENCY DOMAIN, AND POOLED FEATURES.

| Fold | Time domain features (%) | Frequency domain features (%) | Pooled features (%) |
|---|---|---|---|
| 1 | 96.51 | 94.26 | 96.96 |
| 2 | 96.38 | 94.74 | 96.91 |
| 3 | 96.52 | 94.78 | 96.86 |
| 4 | 96.26 | 94.83 | 96.83 |
| 5 | 96.44 | 93.71 | 96.97 |
| 6 | 96.10 | 95.17 | 96.66 |
| 7 | 96.12 | 95.30 | 97.36 |
| 8 | 96.16 | 94.86 | 97.11 |
| 9 | 96.33 | 94.49 | 97.39 |
| 10 | 96.36 | 94.86 | 97.16 |
| Average | 96.32 | 94.70 | 97.02 |

Table IX and Table X provide the confusion matrix for applying RF-SVM in the proposed framework using time domain features and the pooled features, respectively.

TABLE IX
CONFUSION MATRIX FOR RF-SVM USING TIME DOMAIN FEATURES

|  |  | Actual |  |  |  |  |  |
|---|---|---|---|---|---|---|---|
|  |  | Bike | Car | Walk | Run | Bus | Precision |
| Predicted | Bike | 97.83 | 0.75 | 1.32 | 0.72 | 2.02 | 95.39 |
|  | Car | 0.44 | 94.74 | 0.15 | 0.05 | 3.84 | 95.51 |
|  | Walk | 1.03 | 0.10 | 97.61 | 0.98 | 0.15 | 97.80 |
|  | Run | 0.00 | 0.00 | 0.20 | 97.63 | 0.05 | 99.74 |
|  | Bus | 0.69 | 4.41 | 0.73 | 0.62 | 93.93 | 93.50 |
|  | Recall | 97.83 | 94.74 | 97.61 | 97.63 | 93.93 |  |

TABLE X
CONFUSION MATRIX FOR RF-SVM USING POOLED FEATURES

|  |  | Actual |  |  |  |  |  |
|---|---|---|---|---|---|---|---|
|  |  | Bike | Car | Walk | Run | Bus | Precision |
| Predicted | Bike | 96.12 | 0.34 | 1.17 | 0.06 | 0.71 | 97.79 |
|  | Car | 0.69 | 96.81 | 0.16 | 0.01 | 2.85 | 96.27 |
|  | Walk | 1.22 | 0.10 | 97.27 | 0.36 | 0.44 | 97.82 |
|  | Run | 0.65 | 0.05 | 1.18 | 99.54 | 0.48 | 97.55 |
|  | Bus | 1.32 | 2.70 | 0.22 | 0.04 | 95.52 | 95.67 |
|  | Recall | 96.12 | 96.81 | 97.27 | 99.54 | 95.52 |  |

## VII. CONCLUSIONS AND RECOMMENDATIONS FOR FUTURE WORK

This study proposes a two-layer hierarchical framework classifier to distinguish between five transportation modes using new extracted frequency domain features pooled with traditionally used time domain features. The first layer contains a multiclass classifier that discriminates between five transportation modes and identifies the two most probable modes. The second layer consists of binary classifiers that differentiate between the two modes identified in the first layer. The outputs of the two layers are combined using Bayes' rule to choose the transportation mode with the largest posterior probability.

We also investigated the possibility of improving the classification accuracy using pooled features in the proposed framework by applying a number of different classification techniques, including KNN, CART, SVM, RF, and RF-SVM. The results showed that using pooled features in the proposed framework increased the classification accuracy for all of the applied classifiers. For the same data, the highest reported accuracy was 95.10% using the traditional approach for detection, whereas the proposed approach in this study achieved an accuracy of 97.02%. This implies that (a) pooling new features to be selected as classifying features increases the classification accuracy regardless of the applied approach and algorithm, and (b) applying the proposed hierarchal framework further increases the classification accuracy. In summary, the proposed hierarchical framework outperformed the traditional approach of applying only a single layer of classifiers.

Although using pooled features increases the classification accuracy, using the new extracted features alone (i.e., frequency domain) results in a lower accuracy than only using time domain features. Transferring time domain into a new space (i.e., frequency domain) and using the magnitude of the first 20 components enhances the control on the information loss. This means that combining different features together in a big pool and then choosing the best subset of features returns better results than using one domain of features alone. The heterogeneous classifier, using RF in the first layer and SVM in the second layer, was found to produce the best overall performance.

As a future recommendation, deep analysis, such as Canonical Correlation Analysis, should be used to correlate between the features in order to obtain better coordinated results. Furthermore, future work should investigate the sensitivity of the results to the monitoring period and the potential use of GPS data.

## VIII. REFERENCES

[1] M. Susi, V. Renaudin, and G. Lachapelle, "Motion Mode Recognition and Step Detection Algorithms for Mobile Phone Users," *Sensors,* vol. 13, pp. 1539-62, 2013.
[2] M. Elhenawy, A. Jahangiri, and H. A. Rakha, "Smartphone Transportation Mode Recognition using a Hierarchical Machine Learning Classifier," presented at the 23rd ITS World Congress, Melbourne, Australia, 2016.
[3] X. Yu, D. Low, T. Bandara, P. Pathak, L. Hock Beng, D. Goyal, *et al.*, "Transportation activity analysis using smartphones," in *Consumer*

*Communications and Networking Conference (CCNC), 2012 IEEE*, 2012, pp. 60-61.

[4] L. Stenneth, O. Wolfson, P. S. Yu, and B. Xu, "Transportation mode detection using mobile phones and GIS information," in *19th ACM SIGSPATIAL International Conference on Advances in Geographic Information Systems, ACM SIGSPATIAL GIS 2011, November 1, 2011 - November 4, 2011*, Chicago, IL, United States, 2011, pp. 54-63.

[5] A. Jahangiri, H. A. Rakha, and T. A. Dingus, "Developing a system architecture for cyclist violation prediction models incorporating naturalistic cycling data," *Procedia Manufacturing,* vol. 3, pp. 5543-5550, 2015.

[6] A. Jahangiri, H. A. Rakha, and T. A. Dingus, "Adopting Machine Learning Methods to Predict Red-light Running Violations," in *Intelligent Transportation Systems (ITSC), 2015 IEEE 18th International Conference on*, 2015, pp. 650-655.

[7] S. Reddy, M. Mun, J. Burke, D. Estrin, M. Hansen, and M. Srivastava, "Using Mobile Phones to Determine Transportation Modes," *Acm Transactions on Sensor Networks,* vol. 6, Feb 2010.

[8] V. Manzoni, D. Maniloff, K. Kloeckl, and C. Ratti, "Transportation mode identification and real-time CO2 emission estimation using smartphones," Technical report, Massachusetts Institute of Technology, Cambridge, 2010.

[9] M. Susi, V. Renaudin, and G. Lachapelle, "Motion mode recognition and step detection algorithms for mobile phone users," *Sensors,* vol. 13, pp. 1539-1562, 2013.

[10] J. R. Kwapisz, G. M. Weiss, and S. A. Moore, "Activity recognition using cell phone accelerometers," *ACM SigKDD Explorations Newsletter,* vol. 12, pp. 74-82, 2011.

[11] L. Wang, L. Zhang, and Z. Yi, "Trajectory predictor by using recurrent neural networks in visual tracking," *IEEE transactions on cybernetics,* vol. 47, pp. 3172-3183, 2017.

[12] A. Jahangiri and H. A. Rakha, "Applying machine learning techniques to transportation mode recognition using mobile phone sensor data," *IEEE Transactions on Intelligent Transportation Systems,* vol. 16, pp. 2406-2417, 2015.

[13] L. Zhang, M. Qiang, and G. Yang, "Mobility transportation mode detection based on trajectory segment," *Journal of Computational Information Systems,* vol. 9, pp. 3279-3286, 2013.

[14] Y. Zheng, L. Liu, L. Wang, and X. Xie, "Learning transportation mode from raw gps data for geographic applications on the web," presented at the Proceedings of the 17th International Conference on World Wide Web, Beijing, China, 2008.

[15] B. Nham, K. Siangliulue, and S. Yeung, "Predicting mode of transport from iphone accelerometer data," Tech. report, Stanford Univ, 2008.

[16] T. Nick, E. Coersmeier, J. Geldmacher, and J. Goetze, "Classifying means of transportation using mobile sensor data," in *Neural Networks (IJCNN), The 2010 International Joint Conference on*, 2010, pp. 1-6.

[17] A. Bolbol, T. Cheng, I. Tsapakis, and J. Haworth, "Inferring hybrid transportation modes from sparse GPS data using a moving window SVM classification," *Computers, Environment and Urban Systems,* 2012.

[18] P. Widhalm, P. Nitsche, and N. Brandie, "Transport mode detection with realistic Smartphone sensor data," in *2012 21st International Conference on Pattern Recognition (ICPR 2012), 11-15 Nov. 2012*, Piscataway, NJ, USA, 2012, pp. 573-6.

[19] S. Reddy, M. Mun, J. Burke, D. Estrin, M. Hansen, and M. Srivastava, "Using mobile phones to determine transportation modes," *ACM Transactions on Sensor Networks (TOSN),* vol. 6, p. 13, 2010.

[20] F. Biljecki, H. Ledoux, and P. Van Oosterom, "Transportation mode-based segmentation and classification of movement trajectories," *International Journal of Geographical Information Science,* vol. 27, pp. 385-407, 2013.

[21] P. G. Fedor-Freybergh and M. Mikulecký, "From the descriptive towards inferential statistics. Hundred years since conception of the Student's t-distribution," *Neuroendocrinol Lett,* vol. 26, pp. 167-171, 2005.

[22] P.-N. Tan, *Introduction to Data Mining,* India: Pearson Education, 2006.

[23] J. H. Friedman, F. Baskett, and L. J. Shustek, "A relatively efficient algorithm for finding nearest neighbors," *IEEE Trans. Comput.,* vol. 24, pp. 1000-1006, 1974.

[24] L. Olshen and C. J. Stone, "Classification and regression trees," *Wadsworth International Group,* vol. 93, p. 101, 1984.

[25] C.-W. Hsu and C.-J. Lin, "A comparison of methods for multiclass support vector machines," *Neural Networks, IEEE Transactions on,* vol. 13, pp. 415-425, 2002.

[26] L. Breiman, "Random forests," *Machine learning,* vol. 45, pp. 5-32, 2001.

[27] W. Y. Loh, "Classification and regression trees," *Wiley Interdisciplinary Reviews: Data Mining and Knowledge Discovery,* vol. 1, pp. 14-23, 2011.

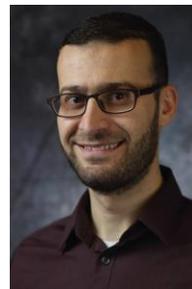
**Huthaifa I. Ashqar** received the B.Sc. degree (with honors) in civil engineering from An-Najah National University, Nablus, Palestine, in 2013 and his M.Sc. in Road Infrastructure from University of Minho, Braga, Portugal, in 2015. He is currently a Ph.D. candidate in the Charles E. Via, Jr. Department of Civil and Environmental Engineering at Virginia Tech, and a trainee at Urban Computing Ph.D. Certificate Program, Discovery Analytics Center, Virginia Tech Blacksburg. He works at the Center for Sustainable Mobility at the Virginia Tech Transportation Institute. His research interests include urban computing, sustainable urban infrastructure, traffic control, traffic modeling and simulation, traffic flow theory, safety modeling, artificial intelligence, and intelligent transportation system.

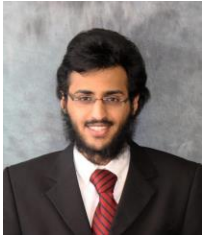
**Mohammed Almannaa** received his B.Sc. in civil engineering with honors: Magna cum Laude from King Saud University, Riyadh, Saudi Arabia, in 2012. He also recently received his M.Sc. in civil engineering from Virginia Polytechnic Institute and State University, Blacksburg, VA, U.S.A, in spring 2016. Currently, he is pursuing his Ph.D. degree, also, in civil engineering at Virginia Polytechnic Institute and State University. He is also a teaching assistant at the Civil Engineering Department at King Saud University. His research interests include but are not limited to eco-driving, highway transportation safety, and intelligent transportation systems, and bike sharing system.

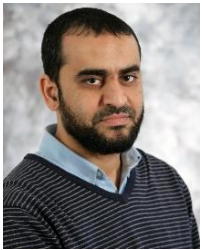
**Mohammed Elhenawy** is a postdoctoral research at Virginia Tech Transportation Institute (VTTI), Blacksburg, Virginia and has been at VTTI since 2015. He received his Ph.D. in computer engineering from Virginia Tech (VT) in 2015. Dr. Elhenawy research interests includes machine learning, statistical learning, game theory and their application in Intelligent Transportation systems (ITS). He has authored or co-authored over 27 ITS related papers.

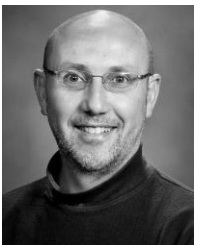
**Hesham A. Rakha** (M'04) received his B.Sc. degree (with honors) in civil engineering from Cairo University, Cairo, Egypt, in 1987 and his M.Sc. and Ph.D. degrees in civil and environmental engineering from Queen's University, Kingston, ON, Canada, in 1990 and 1993, respectively. He is currently the Samuel Reynolds Pritchard Professor of engineering with the Charles E. Via, Jr. Department of Civil and Environmental Engineering at Virginia Tech, a Courtesy Professor with the Bradley Department of Electrical and Computer Engineering at Virginia Tech, Blacksburg, and the Director of the Center for Sustainable Mobility at the Virginia Tech Transportation Institute. He has authored/coauthored 400 refereed publications in the areas of traffic flow theory, traffic modeling and simulation, traveler and driver behavior modeling, artificial intelligence, dynamic traffic assignment, traffic control, energy and environmental modeling, and safety modeling. Dr. Rakha in addition to being a member of IEEE is a member of the ITE, the ASCE, the SAE and TRB. He is a Professional Engineer in Ontario, Canada.

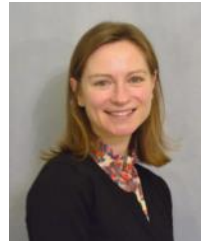
**Leanna House** is an Associate Professor of Statistics at Virginia Tech (VT), Blacksburg, Virginia and has been at VT since 2008. Prior to VT, she worked at Battelle Memorial Institute, Columbus, Ohio; received her Ph.D. in Statistics from Duke University, Durham, North Carolina in 2006; and subsequently served as a post-doctoral research associate for two years in the Department of Mathematical Sciences at Durham University, Durham, United Kingdom. Dr. House has authored or co-authored over 25 journal papers and has been a strong statistical contributor to successful grant proposals including, "NRT-DESE: UrbComp: Data Science for Modeling, Understanding, and Advancing Urban Populations", "Usable Multiple Scale Big Data Analytics Through Interactive Visualization", "Critical Thinking with Data Visualization", "Examining the Taxonomic, Genetic, and Functional Diversity of Amphibian Skin Microbiota", and "Bayesian Analysis and Visual Analytics".